%% file: main.tex
\documentclass[10pt,twocolumn,letterpaper]{article}

\PassOptionsToPackage{table}{xcolor} 

\usepackage{cvpr}              


\definecolor{cvprblue}{rgb}{0.21,0.49,0.74}
\usepackage[pagebackref,breaklinks,colorlinks,allcolors=cvprblue]{hyperref}

\usepackage{multirow}
\usepackage{amsmath}
\usepackage{makecell}
\usepackage{pifont}
\usepackage{bbding}
\usepackage{bm}
\usepackage{adjustbox}
\usepackage{stfloats}

\def\paperID{9960} 
\def\confName{CVPR}
\def\confYear{2025}

\title{MI-DETR: An Object Detection Model with Multi-time Inquiries Mechanism}

\author{
Zhixiong Nan$^{1}$, Xianghong Li$^1$, Jifeng Dai$^2$, Tao Xiang$^{1}$\\
$^1$ College of Computer Science, Chongqing University, Chongqing, China. \\
$^2$ Department of Electronic Engineering, Tsinghua University, Beijing, China. \\
\texttt{nanzx}@cqu.edu.cn,
\texttt{lixianghong}@stu.cqu.edu.cn,
\texttt{txiang}@cqu.edu.cn \\
\texttt{daijifeng}@tsinghua.edu.cn
}

\begin{document}
\maketitle
\begin{abstract}
Based on analyzing the character of cascaded decoder architecture commonly adopted in existing DETR-like models, this paper proposes a new decoder architecture. The cascaded decoder architecture constrains object queries to update in the cascaded direction, only enabling object queries to learn relatively-limited information from image features. However, the challenges for object detection in natural scenes (e.g., extremely-small, heavily-occluded, and confusingly mixed with the background) require an object detection model to fully utilize image features, which motivates us to propose a new decoder architecture with the parallel \textbf{Multi-time Inquiries (MI)} mechanism. \textbf{MI} enables object queries to learn more comprehensive information, and our \textbf{MI} based model, \textbf{MI-DETR}, outperforms all existing DETR-like models on COCO benchmark under different backbones and training epochs,  achieving \textbf{+2.3} AP and \textbf{+0.6} AP improvements compared to the most representative model DINO and SOTA model Relation-DETR under ResNet-50 backbone. In addition, a series of diagnostic and visualization experiments demonstrate the effectiveness, rationality, and interpretability of \textbf{MI}.
\end{abstract}

\section{Introduction}
\label{sec:intro}
The first widely-acknowledged object detection model, Viola Jones detector~\cite{990517}, is proposed in 2001. Fueled by the remarkable R-CNN~\cite{girshick2014rich} and DETR~\cite{carion2020end}, CNN based works~\cite{girshick2015fast,ren2015faster,redmon2016you,lin2017focal,redmon2018yolov3,cai2019cascade,chen2019hybrid,tian2019fcos,liu2016ssd} have significantly pushed forward the object detection technique since 2014, and transformer based DETR-like models~\cite{zhu2020deformable,yao2021efficient,meng2021conditional,wang2022anchor,liu2022dab,li2022dn,zhang2022dino,gao2022adamixer,liu2023detection,cai2023align,zhang2023dense,ye2023cascade,chen2023group,jia2023detrs,liu2023box,hou2024salience,zhao2024ms,hu2024dac,hou2024relation} further bring the technique to a new landscape since 2020. A DETR-like model consists of a backbone, a transformer encoder, a transformer decoder, and a prediction head. From the perspective of function, the backbone and encoder are responsible for image features extraction, the decoder utilizes image features to adapt to the object detection task (\ie, ``feature utilization"), and the prediction head predicts locations and categories of objects.

This paper attempts to explore a new decoder architecture to optimize feature utilization.
Each transformer decoder layer contains a self-attention (SA), a cross-attention (CA), and a feed-forward network (FFN), where object queries are optimized by interacting with image features.  Analogously speaking, it is like that a student (\textit{object queries act like a student who ``knows'' nothing or a little about the image at the beginning decoder layer}) asks a question to a teacher (\textit{image features could be analogous to a teacher since he ``knows'' all information in the image}) and obtains the image information based on the teacher's answer. Therefore, the ``SA+CA+FFN'' architecture in each decoder layer is analogously termed as ``inquiry head" in this paper.
The transformer decoder of existing DETR-like models adopts the  ``cascaded inquiries" architecture. Object queries continuously inquire the image features layer by layer to learn the gradually-refined information in the cascaded direction.

The robustness of this decoder architecture has been fully validated in previous DETR-like models.
However, from the perspective of feature utilization, this architecture presents a worth-noting character. The representations of object queries in the next decoder layer are directly decided by that in the current decoder layer, thus the update of query representations is constrained in the cascaded direction, which indicates that the cascaded inquiries architecture tends to learn the relatively limited information.
In addition, excessively refined information learnt in deep decoder layers of the same cascaded direction might be redundant or even negative. More analysis could be found in experiment and discussion sections.

\begin{figure}[t]
            \centering
            \includegraphics[width=0.48\textwidth]{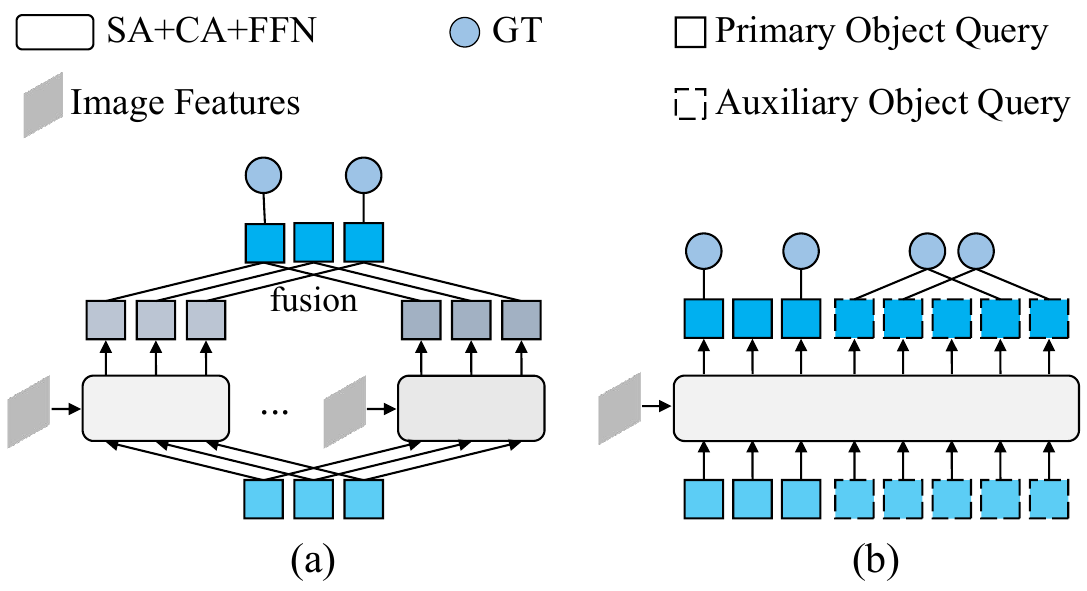}
            \caption{(a) We propose the parallel multi-time inquiries mechanism with parameters-dependent inquiry heads and fusion operation in a decoder layer; (b) In recent proposed DETR-like models, primary and auxiliary queries are concatenated and inputted to the same inquiry head in a decoder layer, thus this kind of parallel architecture is parameters-sharing parallel. 
            }
            \label{fig:Intro}
            \vspace{-3mm}
\end{figure}

Due to the diversity and complexity of natural scenes, objects might be extremely-small, heavily-occluded, or confusingly mixed with the background.
To adapt to diverse scenes, it is required to fully utilize image features to learn comprehensive information.
However, based on the above analysis, the cascaded multi-time inquiries architecture adopted in existing models tends to learn the relatively limited information. This motivates us to explore a new decoder architecture to optimize feature utilization.
Inspired by the development of traditional CNN-based methods, we have noticed that many methods~\cite{zagoruyko2016wide,szegedy2015going,szegedy2016rethinking,szegedy2017inception,li2019scale} enhance feature utilization through reasonable parallel architectures. Therefore, we propose a new decoder architecture with the parallel \textbf{\textit{Multi-time Inquiries (MI) }}mechanism.
As shown in \cref{fig:Intro}{\textcolor[RGB]{54,125,189}{a}}, the architecture is simple, object queries parallelly perform multi-time inquiries through multiple parameters-independent inquiry heads, and the object queries are then fused. The parallel \textbf{\textit{MI}} mechanism conducts feature utilization in separate branches, allowing to learn gradually-refined multi-pattern information, and these information can be fused to be comprehensive.
Apart from \textbf{\textit{MI}} mechanism, we also design a \textbf{\textit{U-like Feature Interaction (UFI)}} module to further improve feature utilization.

It is noted that the proposed decoder architecture is distinct from the parameters-sharing parallel decoder adopted in recent proposed DETR-like models, such as $\mathcal{H}$-DETR~\cite{jia2023detrs} and Group-DETR~\cite{chen2023group}, whose decoder architecture is illustrated in \cref{fig:Intro}{\textcolor[RGB]{54,125,189}{b}}. The differences are mainly reflected on two aspects.
\textbf{\textit{First}}, from the perspective of motivation, the architecture in \cref{fig:Intro}{\textcolor[RGB]{54,125,189}{b}} is designed to solve the issue of insufficient positive object queries (\ie the object queries that are supervised by Ground Truths), thus auxiliary object queries and one-to-many supervision are introduced.
Differently, our method is proposed to optimize feature utilization by digging multi-pattern information.
\textbf{\textit{Second}}, from the perspective of architecture, primary and auxiliary object queries in \cref{fig:Intro}{\textcolor[RGB]{54,125,189}{b}} are concatenated together and then inputted into the same inquiry head, thus the parameters are sharing and simultaneously-updating.
In our opinion, this type of parameters-sharing parallel architecture, from the perspective of feature utilization, is the pseudo parallel architecture, since sharing parameters make two kinds of queries learn the same pattern of information. In contrast, our parameters-independent parallel architecture allows object queries to learn multi-pattern information to improve feature utilization. More analysis could be found in the discussion section.

The proposed decoder architecture with the parallel \textbf{\textit{MI}} mechanism is simple and friendly, and it can be easily plugged into existing DETR-like models. In the experiments, by plugging into the most representative model DINO~\cite{zhang2022dino} and SOTA Relation-DETR~\cite{hou2024relation}, impressive performance improvements are obtained, achieving \textbf{+2.3} AP and \textbf{+0.6} AP improvements, respectively, under ResNet-50 backbone and the condition achieving convergence. To the best of our knowledge, these are the best results among existing DETR-like models.

The contributions are as follows: \textit{\textbf{i)}} This paper proposes a new decoder architecture to optimize feature utilization by digging multi-pattern information, and our \textbf{MI-DETR} (\ie, \textbf{Multi-time Inquiries DETR}) achieves the best performance among existing DETR-like models; \textit{\textbf{ii)}} The proposed decoder architecture is simple and friendly to plug into existing DETR-like models. The code will be released.

\section{Related Work}
\label{sec:releted_work}
Original DETR~\cite{carion2020end} suffers from slow convergence, requiring 500 epochs to achieve 43.3 AP. In recent years, a large amount of works have focused on improving detection performance and accelerating training convergence of DETR, bringing the significant performance breakthrough to 51.7 AP within 12 epochs.

In the early stage, most methods focus on optimizing the representations of object queries. Object queries are learnable vectors that are used to capture information about objects in the image. Some works~\cite{zhu2020deformable,yao2021efficient,zhang2022dino,li2023mask,zhao2024detrs,hou2024salience} have noticed that the design of the initial object queries will significantly impact the convergence speed and propose two-stage initialization to select top K encoder features from the last encoder layer to enhance the initial object queries. Furthermore, many studies~\cite{meng2021conditional,liu2022dab,liu2023box,zhang2024enhancing,huang2025dq} consider introducing prior knowledge into object queries. In addition, the attention mechanism is also a key area of interest in early researches. Cross-attention performs the interaction between object queries and image features, which enables the object queries to concentrate on related regions of the image. Several methods~\cite{zhu2020deformable,dai2021dynamic,gao2021fast,wang2022anchor,ye2023cascade} have optimized the cross-attention in decoder to improve the ability of object queries to quickly locate relevant regions.

In the past two years, many researchers have paid attention to the limitations of one-to-one supervision on convergence speed and performance. A few recent studies~\cite{chen2023group,jia2023detrs,zhao2024ms,ouyang2022nms,wang2022towards} have proposed some effective one-to-many matching strategies, allowing a GT to match with multiple object queries to accelerate training speed.

\begin{figure*}[t]
            \centering
            \includegraphics[width=1.0\textwidth]{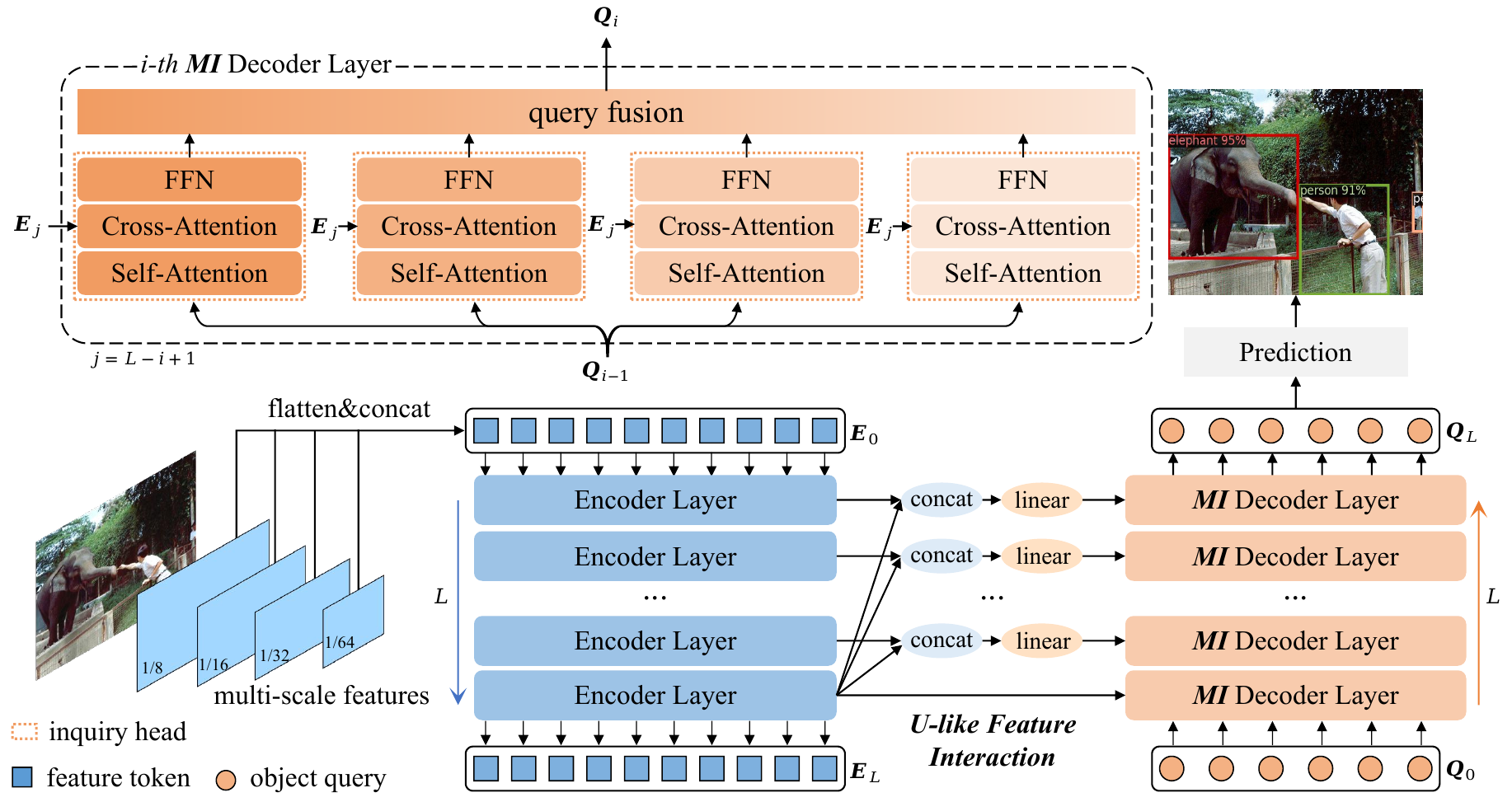}
            \caption{The overview of \textbf{MI-DETR}.
            The main novelty is that \textbf{MI-DETR} uses \textbf{\textit{Multi-time Inquiries (MI)}} decoder layers to replace the traditional decoder layers adopted in previous DETR-like models.
            Backbone and $L$-layer transformer encoder extract the image features $\bm{E}=\{\bm{E}_0,  \ldots, \bm{E}_L\}$.
            For $i$-\textit{th} \textbf{\textit{MI}} decoder layer, the input is object queries $\bm{Q}_{i-1}$ and the output is $\bm{Q}_i$.
            For inquiry heads in $i$-\textit{th} \textbf{\textit{MI}} decoder layer, object queries learn multi-pattern information by interacting with $\bm{E}_j$, where $j=L-i+1$, and $\bm{E}_j$ is the corresponding image features after the processing of \textbf{\textit{U-like Feature Interaction}}.
            The output of the last \textbf{\textit{MI}} decoder layer (\ie $\bm{Q}_L$) is used to predict the locations and categories of objects.
             }
            \label{fig:overview}
            \vspace{-2mm}
\end{figure*}

\section{Method}
\label{sec:methods}

\subsection{Preliminaries}
\textbf{The framework of DETR-like models.} DETR-like models usually present the similar framework with the original DETR~\cite{carion2020end}.
\textit{First}, an image is inputted to the backbone to obtain multi-scale image features, which are flattened into feature tokens, and then concatenated to serve as the input of transformer encoder to refine image features. \textit{Second}, a group of learnable object queries (object candidates are represented in the form of object queries~\cite{zhao2024ms}) interact with image features in transformer decoder to obtain the refined object queries. \textit{Third}, the refined object queries are used to predict bounding boxes and categories of objects.

\subsection{U-like Feature Interaction}
\label{sec:UFI}
In existing DETR-like models, although the backbone and transformer encoder have effectively extracted image features, these features have not been sufficiently utilized. Specifically, only the feature at the last layer of the transformer encoder is leveraged as Key\&Value for transformer decoder. Actually, the features from each layer of transformer encoder contain valuable information at different levels.
Inspired by the classical U-Net~\cite{ronneberger2015u}, we propose \textbf{\textit{U-like Feature Interaction (UFI)}} to fully make use of features at different transformer encoder layers.

DETR is a typical encoder-decoder architecture.
Overall, the learning of DETR is a process of encoding low-level detailed features to high-level abstract features, and then decoding them back into detailed representations.
The encoder extracts image features layer by layer. With the number of encoder layers increasing, the extracted features become more global and abstract. The decoder continuously refines the object queries to gradually learn the local and detailed information. Therefore, the interaction between the corresponding layers of the encoder and decoder facilitates to utilize both low-level and high-level information.

The image features at encoder layers are defined as $\bm{E}=\{\bm{E}_j, j=1, 2, \ldots, L\}$, where $L$ denotes the number of encoder layers. \textbf{\textit{UFI}} uses the features at the $j$-\textit{th} layer of the transformer encoder (\ie, $\bm{E}_j$) as Key\&Value for the $i$-\textit{th} decoder layer, where $j=L-i+1$.
In detail, \textbf{\textit{UFI}} first fuses the features at the last layer of the transformer encoder ($\bm{E}_L$) with the features at other layers $\{\bm{E}_1, \bm{E}_2, ... , \bm{E}_{L-1}\}$, 
\begin{small}
\begin{equation}
\label{eq:Ej}
\bm{E}_j\! =\!
\begin{cases}
    \operatorname{linear}(\operatorname{concat}(\bm{E}_{j}, \bm{E}_{L})) &j\! =\! 1,2, \ldots, L-1 \\
    \bm{E}_{L} & j\! =\! L
\end{cases},
\end{equation}
\end{small}
and then uses these fused features as the Key\&Value of the corresponding decoder layers. The $i$-\textit{th} decoder layer could be formulated as follows: 
\begin{small}
\begin{equation}
    \bm{Q}_{i}\! =\! \mathcal{D}_{i}(\bm{Q}_{i-1}, \bm{E}_{j}),\  j\! =\! L-i+1,
\end{equation}
\end{small}
where $\mathcal{D}_{i}$ represents $i$-\textit{th} decoder layer.

\subsection{Multi-time Inquiries Mechanism}
\label{sec:PMID}
The key idea of parallel \textbf{\textit{Multi-time Inquiries}} (\textbf{\textit{MI}}) mechanism is to make object queries perform multiple interactions with image features to improve feature utilization. Specifically, we apply \textbf{\textit{MI}} mechanism to original transformer decoder layers, yielding \textbf{\textit{MI Decoder}}. Each layer can be divided two parts: \textit{\textbf{multi-time inquiries}} and \textbf{\textit{query fusion}}.

\textbf{Multi-time inquiries.} As shown in \cref{fig:overview}, the inputs of the $i$-\textit{th} \textbf{\textit{MI}} decoder layer are the object queries $\bm{Q}_{i-1}$ (\ie, the output of $(i\text{-1})$-\textit{th} \textbf{\textit{MI}} decoder layer) and the corresponding image features $\bm{E}_j$ defined in \cref{eq:Ej}. We process the object queries $\bm{Q}_{i-1}$ with $M$ different inquiry heads. Same with the network architecture of traditional transformer decoder layer, each inquiry head is composed of a self-attention layer, a cross-attention layer, and a FFN layer, which is formalized as follows:
\begin{small}
\begin{equation}
\label{eq: PMI}
\begin{array}{cc}
     & \bm{Q}_{i}^k\! =\! \operatorname{FFN}_{i}^k(\operatorname{CrossAtt}_{i}^k(\operatorname{SelfAtt}_{i}^k(\bm{Q}_{i-1}), \bm{E}_j)), \\
     & \\
     & i\! =\! 1, 2, \ldots, L; k\! =\! 1, 2, \ldots, M,
\end{array}
\end{equation}
\end{small}
where $\bm{Q}_{i}^k$ denotes the object queries outputted from $k$-\textit{th} inquiry head at $i$-\textit{th} \textbf{\textit{MI}} decoder layer. $\operatorname{SelfAtt}_{i}^k$, $\operatorname{CrossAtt}_{i}^k$, and $\operatorname{FFN}_{i}^k$ represent the self-attention layer, cross-attention layer, and FFN layer in $k$-\textit{th} inquiry head of $i$-\textit{th} \textbf{\textit{MI}} decoder layer. $L$ is the number of decoder layers.

\textbf{Query fusion.} 
Based on \cref{eq: PMI}, $M$ groups of object queries $\{\bm{Q}_{i}^1, \bm{Q}_{i}^2, \ldots, \bm{Q}_{i}^M\}$ are computed. Different groups of object queries convey different patterns of information. Therefore, it is necessary to conduct fusion to make them to be mutually cooperative and beneficial. We adopt the classic concatenation fusion, which first concatenates multiple groups of object queries along the feature dimension and then projects the concatenated feature to the original dimensions.
\begin{small}
\begin{equation}
\label{eq:fusion}
    \bm{Q}_{i}\! =\! \operatorname{Linear}(\operatorname{Concat}(\bm{Q}_{i}^1, \bm{Q}_{i}^2, \ldots, \bm{Q}_{i}^M)),
\end{equation}
\end{small}
where $\operatorname{Concat}$ denotes concatenation operation and $\operatorname{Linear}$  represents a linear layer.

\begin{figure}[h]
            \centering
            \includegraphics[width=0.48\textwidth]{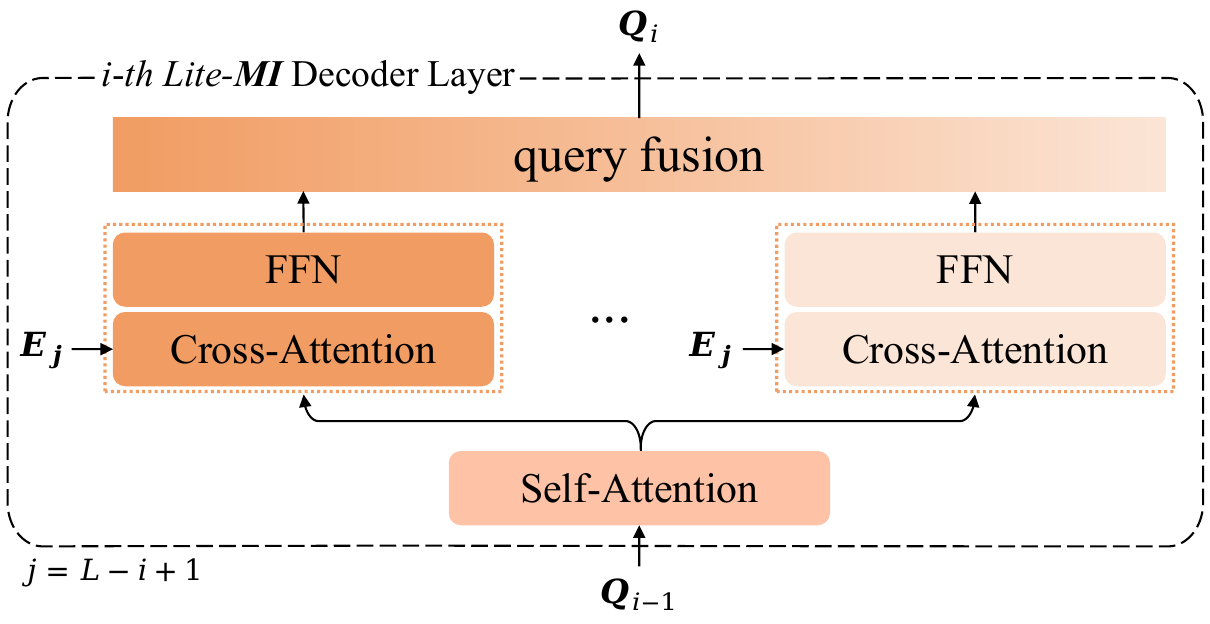}
               \caption{The architecture of \textit{Lite-\textbf{MI}}. 
               }
            \label{fig:PMID++}
            \vspace{-2mm}
\end{figure}

\subsection{Lite Multi-time Inquiries}
\textit{Lite} \textbf{\textit{Multi-time Inquiries}} (\textit{Lite}-\textbf{\textit{MI}}) is the lightweight version of \textbf{\textit{MI}}. As shown in \cref{fig:PMID++}, the parameters of self-attention layers in different inquiry heads are shared, which is formulated as follows:
\begin{small}
\begin{equation}
\begin{array}{cc}
     & \bm{Q}_{i}^k\! =\! \operatorname{FFN}_{i}^k(\operatorname{CrossAtt}_{i}^k(\operatorname{SelfAtt}_{i}(\bm{Q}_{i-1}), \bm{E}_j)), \\
\end{array}
\end{equation}
\end{small}
where $\operatorname{SelfAtt}_{i}$ represents the shared self-attention layer.

The motivations of \textit{Lite}-\textbf{\textit{MI}} are two fold. \textit{First}, as studied in~\cite{hu2024dac}, the main function of self-attention is to eliminate duplicate candidates. In addition, image features $\bm{E}_j$ are not involved in the self-attention layer. Therefore, configuring separate self-attention layers in all inquiry heads might be redundant. \textit{Second}, it is able to reduce parameters.

\section{Experiments}

\begin{table*}[t]
\centering
\scalebox{0.85}{
        \setlength{\tabcolsep}{3.2mm}{
        \renewcommand\arraystretch{1.0}
                \begin{tabular}{llcccccccc}
                        \toprule[1.0pt]
                        Method   & Pub.Year & Backbone & Epochs & AP  & $AP_{50}$ & $AP_{75}$ & $AP_{S}$ & $AP_{M}$ & $AP_{L}$ \\
                        \midrule[1.0pt]
                        DINO~\cite{zhang2022dino} & ICLR.2023 & ResNet-50 & 12  & 49.0 & 66.6 & 53.5 & 32.0 & 52.3 & 63.0 \\
                        Stable-DINO~\cite{liu2023detection} & ICCV.2023 & ResNet-50 & 12  & 50.4 & 67.4 & 55.0 & 32.9 & 54.0 & 65.5 \\
                        Rank-DETR~\cite{pu2024rank} & NIPS.2023 & ResNet-50 & 12  & 50.4 & 67.9 & 55.2 & 33.6 & 53.8 & 64.2 \\
                        Align-DETR~\cite{cai2023align} & BMVC.2024 & ResNet-50 & 12  & 50.2 & 67.8 & 54.4 & 32.9 & 53.3 & 65.0 \\
                        DDQ-DETR~\cite{zhang2023dense} & CVPR.2023 & ResNet-50 & 12  & 51.3 & 68.6 & 56.4 & 33.5 &54.9 &65.9    \\
                        DAC-DETR~\cite{hu2024dac} & NIPS.2023 & ResNet-50 & 12  & 50.0 & 67.6 & 54.7 & 32.9 & 53.1 & 64.2 \\
                        $\mathcal{H}$-DETR~\cite{jia2023detrs} & CVPR.2023 & ResNet-50 & 12  & 48.7 & 66.4 & 52.9 & 31.2 & 51.5 & 63.5 \\
                        Group-DETR~\cite{chen2023group} & ICCV.2023 & ResNet-50 & 12  & 49.8 & - & - & 32.4 & 53.0 & 64.2 \\
                        MS-DETR~\cite{zhao2024ms} & CVPR.2024 & ResNet-50 & 12  & 50.3 & 67.4 & 55.1 & 32.7 & 54.0 & 64.6 \\
                        Relation-DETR~\cite{hou2024relation} & ECCV.2024 & ResNet-50 & 12  & 51.7 & 69.1 & 56.3 & 36.1 & 55.6 & 66.1 \\
                        \hline
                        \rowcolor{lightgray!30} Ours  & -  & ResNet-50 & 12  & \textbf{52.4} & \textbf{69.8} & \textbf{57.0} & 35.6 & \textbf{56.1} & \textbf{67.2} \\
                        \midrule[1.0pt]
                        DINO~\cite{zhang2022dino} & ICLR.2023 & ResNet-50 & 36  & 50.9 & 69.0 & 55.3 & 34.6 & 54.1 & 64.6 \\
                        Stable-DINO~\cite{liu2023detection} & ICCV.2023 & ResNet-50 & 24  & 51.5 & 68.5 & 56.3 & 35.2 & 54.7 & 66.5 \\
                        Rank-DETR~\cite{pu2024rank} & NIPS.2023 & ResNet-50 & 36  & 51.2 & 68.9 & 56.2 & 34.5 & 54.9 & 64.9 \\
                        $\mathcal{H}$-DETR~\cite{jia2023detrs} & CVPR.2023 & ResNet-50 & 36  & 50.0 & 68.3 & 54.4 & 32.9 & 52.7 & 65.0 \\
                        Group-DETR~\cite{chen2023group} & ICCV.2023 & ResNet-50 & 36  & 51.3 & - & - & 34.7 & 54.5 & 65.3 \\
                        Align-DETR~\cite{cai2023align} & BMVC.2024 & ResNet-50 & 24  & 51.3 & 68.2 & 56.1 & 35.5 & 55.1 & 65.6 \\
                        DDQ-DETR~\cite{zhang2023dense} & CVPR.2023 & ResNet-50 & 24  & 52.0 & 69.5 & 56.2 & 35.2 & 54.9 & 65.9 \\
                        DAC-DETR~\cite{hu2024dac} & NIPS.2023 & ResNet-50 & 24  & 51.2 & 68.9 & 56.0 & 34.0 & 54.6 & 65.4 \\
                        MS-DETR~\cite{zhao2024ms} & CVPR.2024 & ResNet-50 & 24  & 51.7 & 68.7 & 56.5 & 34.0 & 55.4 & 65.5 \\
                        Relation-DETR~\cite{hou2024relation} & ECCV.2024 & ResNet-50 & 24  & 52.1 & 69.7 & 56.6 & 36.1 & 56.0 & 66.5 \\
                        \hline
                        \rowcolor{lightgray!30} Ours & - & ResNet-50 & 24  & \textbf{52.7} & \textbf{70.4} & \textbf{57.2} & \textbf{36.7} & \textbf{56.7} & \textbf{66.7} \\
                        \midrule[1.0pt]
                        DINO~\cite{zhang2022dino} & ICLR.2023 & Swin-L & 12  & 56.8 & 75.6 & 62.0 & 40.0 & 60.5 & 73.2 \\
                        DAC-DETR~\cite{hu2024dac} & CVPR.2023 & Swin-L & 12  & 57.3 & 75.7 & 62.7 & 40.1 & 61.5 & 74.4 \\
                        Rank-DETR~\cite{pu2024rank} & NIPS.2023 & Swin-L & 12  & 57.6 & 76.0 & 63.4 & 41.6 & 61.4 & 73.8 \\
                        $\mathcal{H}$-DETR~\cite{jia2023detrs} & CVPR.2023 & Swin-L & 12  & 56.1 & 75.2 & 61.3 & 39.3 & 60.4 & 72.4 \\
                        Stable-DINO~\cite{liu2023detection} & ICCV.2023 & Swin-L & 12  & 57.7 & 75.7 & 63.4 & 39.8 & 62.0 & 74.7 \\
                        Relation-DETR~\cite{hou2024relation} & ECCV.2024 & Swin-L & 12  & 57.8 & 76.1 & 62.9 & 41.2 & 62.1 & 74.4 \\
                        \hline
                        \rowcolor{lightgray!30} Ours  & - & Swin-L  & 12  &\textbf{58.2}  &\textbf{76.5} &\textbf{63.4} &\textbf{42.5}  &\textbf{62.8}  &\textbf{74.6}  \\
                         \bottomrule[1.0pt]
                \end{tabular}
        }}
\caption{Comparison experiments under different backbones and training epochs.}
\vspace{-2mm}
\label{tab:sota-resnet}
\end{table*}

\subsection{Settings}
We conduct experiments on COCO~\cite{lin2014microsoft} dataset. Following the common practice, COCO train2017 split (118k images) is for training and COCO val2017 split (5k images) is for validation. AdamW~\cite{loshchilov2017decoupled} is used for optimization, with the learning rate of $1\times 10^{-4}$ and weight decay of $1\times 10^{-4}$. We report the experimental results under the 1x (12 epochs) and 2x (24 epochs) training schedules with two commonly-used backbones, including ResNet-50~\cite{he2016deep} pretrained on ImageNet-1k~\cite{russakovsky2015imagenet} and Swin-L~\cite{liu2021swin} pretrained on ImageNet-22k. The metric is the standard mean Average Precision (AP) under different IoU thresholds and object scales. RTX3090 GPUs are used when the backbone is ResNet-50 and A100 GPUs are used when the backbone is Swin-L. The number of inquiry heads is 4.

\subsection{Comparison Experiments}
For comprehensive comparison, our model is compared with a series of DETR variants~\cite{zhang2022dino,liu2023detection,pu2024rank,cai2023align,zhang2023dense,hu2024dac,jia2023detrs,chen2023group,zhao2024ms,hou2024relation} that are proposed since 2023, under different training schedules and backbones. The comparison results are summarized in \cref{tab:sota-resnet}. 

\textbf{Comparison under ResNet-50.} With the rapid development of DETR variants, the convergence speed has significantly accelerated. Most methods can achieve convergence within 24 epochs, and a few methods (\eg, DINO~\cite{zhang2022dino}, Rank-DETR~\cite{pu2024rank}, $\mathcal{H}$-DETR~\cite{jia2023detrs}, and Group-DETR~\cite{chen2023group}) require 36 epochs to reach convergence. Therefore, we report experimental results under 12 training epochs (commonly adopted in existing methods) and the condition achieving convergence (\ie, 24 or 36 epochs, depending on the model itself). As shown in \cref{tab:sota-resnet}, our method demonstrates the advantage under different training conditions. Compared to the second-best method (\ie, Relation-DETR ~\cite{hou2024relation}), our method achieves the improvements of \textbf{+0.7} AP and \textbf{+0.6} AP with 12 and 24 training epochs, respectively. \textit{Without bells and whistles, our method establishes the best result to date}, \textit{achieving 52.7 AP without employing extra tricks.}

\textbf{Comparison under Swin-L.} As a stronger backbone, Swin-L backbone has demonstrated its excellent performance. However, few methods report the results under Swin-L backbone. To further verify the effectiveness and robustness of our method, the experiments are conducted using Swin-L backbone under the training schedule of 12 epochs, and the results are reported in \cref{tab:sota-resnet}, from which we can observe that \textit{our method still achieves the best results, obtaining \textbf{+0.4} AP compared with the existing best-performing model Relation-DETR.}

\begin{table}[h!]
\vspace{1mm}
\centering
\scalebox{0.85}{
        \setlength{\tabcolsep}{3mm}{
        \renewcommand\arraystretch{1.0}
                \begin{tabular}{c|ccc|ccc}
                        \toprule[1.0pt]
                        ID & \textbf{\textit{MI}} & \textit{Lite-}\textbf{\textit{MI}} & \textbf{\textit{UFI}} & AP & $AP_{50}$ & $AP_{75}$ \\
                         \midrule[1.0pt]
                        \#1   &- &- &-  & 49.0 & 66.6 & 53.5 \\
                        \cmidrule(lr){1-7}
                        \#2     & \Checkmark &   &   & 49.8 & 67.5 & 54.3 \\
                        \#3     &   & \Checkmark & & 49.6 & 67.2 & 54.1 \\
                        \#4     & &   & \Checkmark & 49.5 & 67.3 & 54.0 \\
                        \#5     &   & \Checkmark & \Checkmark  & 50.1 & 67.9 & 54.7 \\
                        \#6     & \Checkmark &  & \Checkmark & 50.2 & 68.1 & 54.8 \\
                        \bottomrule[1.0pt]
                \end{tabular}
        }}
\caption{Diagnostic experiments on \textbf{\textit{MI}} (\textbf{\textit{Multi-time Inquiries}}), \textit{Lite-\textbf{MI}} (\textit{Lite} \textbf{\textit{Multi-time Inquiries}}), and \textbf{\textit{UFI}} (\textbf{\textit{U-like Feature Interaction}}).
The experiments are conducted based on DINO baseline ($\text{\#1}$), and we adopt ResNet-50 backbone and 1x (12 epochs) training schedule. \textbf{The results in \cref{tab:ablation-numbranch}, \cref{tab:ablation-branch}, and \cref{tab:ablation-layerbranch} are also obtained under the same experiment settings.}}
\vspace{-4mm}
\label{tab:ablation-components}
\end{table}

\begin{table*}[t]
\centering
\scalebox{0.85}{
        \setlength{\tabcolsep}{1.5mm}{
        \renewcommand\arraystretch{1.0}
                \begin{tabular}{llclllllll}
                        \toprule[1.0pt]
                        Method & Backbone & Epochs & AP   & $AP_{50}$ & $AP_{75}$ & $AP_{S}$ & $AP_{M}$ & $AP_{L}$ \\
                        \midrule[1.0pt]
                        DINO~\cite{zhang2022dino} & ResNet-50 & 12  & 49.0 & 66.6   & 53.5   & 32.0  & 52.3  & 63.0     \\
                        \rowcolor{lightgray!30} ours & ResNet-50 & 12  & \textbf{50.2(+1.2)} & \textbf{68.1(+1.5)} & \textbf{54.8(+1.3)} & \textbf{33.4(+1.4)} & \textbf{53.6(+1.3)} & \textbf{64.5(+1.5)} \\
                        DINO~\cite{zhang2022dino}  & ResNet-50 & 24  & 50.4 & 68.3   & 54.8   & 33.3  & 53.7  & 64.8     \\
                        \rowcolor{lightgray!30} ours & ResNet-50 & 24  & \textbf{51.2(+0.8)} & \textbf{69.2(+0.9)} & \textbf{55.8(+1.0)} & \textbf{34.6(+1.3)} & \textbf{54.4(+0.7)} & \textbf{65.5(+0.7)} \\
                        DINO~\cite{zhang2022dino}  & Swin-L & 12  & 56.8 & 75.6   & 62.0   & 40.0  & 60.5  & 73.2     \\
                        \rowcolor{lightgray!30} ours & Swin-L & 12  &\textbf{57.5(+0.7)} &\textbf{76.3(+0.7)}  &\textbf{63.1(+1.1)}  &\textbf{41.1(+1.1)}  &\textbf{61.5(+1.0)}  &\textbf{73.9(+0.7)}        \\
                        \midrule[1.0pt]
                        Relation-DETR~\cite{hou2024relation}  & ResNet-50 & 12  & 51.7 & 69.1   & 56.3   & 36.1  & 55.6  & 66.1     \\
                        \rowcolor{lightgray!30} ours & ResNet-50 & 12  & \textbf{52.4(+0.7)} & \textbf{69.8(+0.7)} & \textbf{57.0(+0.7)} & 35.6 & \textbf{56.1(+0.5)} & \textbf{67.2(+1.1)} \\
                        Relation-DETR~\cite{hou2024relation}  & ResNet-50 & 24  & 52.1 & 69.7   & 56.6   & 36.1  & 56.0  & 66.5     \\
                        \rowcolor{lightgray!30} ours & ResNet-50 & 24  & \textbf{52.7(+0.6)} & \textbf{70.4(+0.7)} & \textbf{57.2(+0.6)} & \textbf{36.7(+0.6)} & \textbf{56.7(+0.7)} & \textbf{66.7(+0.2)} \\
                        Relation-DETR~\cite{hou2024relation}  & Swin-L & 12  & 57.8 & 76.1   & 62.9   & 41.2  & 62.1  & 74.4     \\
                        \rowcolor{lightgray!30} ours & Swin-L & 12  &\textbf{58.2(+0.4)}  &\textbf{76.5(+0.4)} &\textbf{63.4(+0.5)} &\textbf{42.5(+1.3)}  &\textbf{62.8(+0.7)}  &\textbf{74.6(+0.2)}  \\
                        \bottomrule[1.0pt]
                \end{tabular}
        }}
\caption{Diagnostic experiments of \textbf{\textit{MI}} on representative models.}
\vspace{-2mm}
\label{tab:sota-combination}
\end{table*}

\subsection{Diagnostic Experiments}
\subsubsection{The ablation testing on main components}
We conduct a series of diagnostic experiments on main components, including \textbf{\textit{MI}}, \textit{Lite-}\textbf{\textit{MI}}, and \textbf{\textit{UFI}}. The results are summarized in \cref{tab:ablation-components}, from which we can observe that all components exhibit improvements over baseline (+0.8 AP, +0.6 AP, and +0.5 AP, respectively), demonstrating that each component is effective (comparing $\text{\#2}$, $\text{\#3}$, and $\text{\#4}$ with $\text{\#1}$). The combinations ``\textbf{\textit{MI}}+\textbf{\textit{UFI}}" ($\text{\#5}$) and ``\textit{Lite-}\textbf{\textit{MI}}+\textbf{\textit{UFI}}" ($\text{\#6}$) represent two kinds of structures of our model.  ``\textbf{\textit{MI}}+\textbf{\textit{UFI}}" presents the best result, indicating the effectiveness of our architecture.  \textit{Lite-}\textbf{\textit{MI}} reduces the model complexity by sharing the self-attention layer. Compared to ``\textbf{\textit{MI}}+\textbf{\textit{UFI}}", ``\textit{Lite-}\textbf{\textit{MI}}+\textbf{\textit{UFI}}" asks for less parameters at the cost of slight performance degradation.

\subsubsection{The effects of \textit{\textbf{MI}} on other models}
\label{experiments: combination}
\textbf{\textit{MI}} is the core mechanism of our model.
Thanks to its simplification, \textbf{\textit{MI}} could be friendly plugged into existing DETR-like models. To verify the effectiveness, we examine the performance improvements after plugging \textbf{\textit{MI}} into DINO~\cite{zhang2022dino} and Relation-DETR~\cite{hou2024relation}. DINO is the most representative model that is widely known in the object detection domain, and Relation-DETR is the newest model presenting SOTA performance. Therefore, we select them as baselines. To avoid the randomness of experiment results, we conduct the experiments under different backbones and training schedules, and the results are presented in \cref{tab:sota-combination}.

We can observe from \cref{tab:sota-combination} that: \textbf{1)} Our method achieves consistent improvements over baselines. Especially, based on the strongest baseline Relation-DETR leveraging one-to-many matching in the decoder, our method can still obtain the performance improvement, surpassing Relation-DETR by \textbf{+0.7} AP (12 epochs) and \textbf{+0.6} AP (24 epochs). This demonstrates \textbf{\textit{MI}} can effectively strengthen the representations of object queries, even though the object queries have been strengthened by the one-to-many matching mechanism; \textbf{2)} Plugging \textbf{\textit{MI}} into Relation-DETR under the training schedule of \textbf{12} epochs obtains better performance than that of Relation-DETR trained with \textbf{24} epochs (52.4 AP~v.s.~52.1 AP), demonstrating our method can accelerate the training convergence.

\subsubsection{Inquiry head number of \textit{\textbf{MI}}}
We are interested in how will the number of inquiry heads in \textbf{\textit{MI}} decoder layers affect the final performance.
Intuitively, the performance will boost with the increasing of inquiry heads, since multi-time inquiries enable the object queries to learn multi-pattern information. With the further increasing of inquiry heads, the performance should drop, since too many patterns of information might be disturbing or even antagonistic.
Therefore, we conduct the experiments to examine the effectiveness of inquiry head number on model performance, and the results are available in \cref{tab:ablation-numbranch}. When inquiry head number increases from 1 to 4, the performance improves by +0.7 AP, validating the effectiveness of multiple inquiry heads. However, the performance drops when IHN=5, potentially proving our guess that too many patterns of information might be disturbing or even antagonistic. 

\begin{table}[h]
\vspace{1mm}
\centering
\scalebox{0.85}{
        \setlength{\tabcolsep}{3mm}{
        \renewcommand\arraystretch{1.0}
                \begin{tabular}{c|ccccc}
                        \toprule[1.0pt]
                        IHN & 1 & 2 & 3 & 4 & 5\\
                        \midrule[1.0pt]
                        AP & 49.5 & 49.6 & 49.9 & \textbf{50.2} & 49.9 \\
                        \bottomrule[1.0pt]
                \end{tabular}
        }}
\caption{Diagnostic experiments on the number of inquiry heads. ``IHN" is short for ``Inquiry Head Number".}
\vspace{-2mm}
\label{tab:ablation-numbranch}
\end{table}

\begin{table}[h]
\centering
\scalebox{0.85}{
        \setlength{\tabcolsep}{2mm}{
        \renewcommand\arraystretch{1.0}
                \begin{tabular}{c|c|c|ccc}
                        \toprule[1.0pt]
                        Method & IHN & Heads & AP & $AP_{50}$ & $AP_{75}$  \\
                        \midrule[1.0pt]
                        DINO  & - & - & 49.0 & 66.6 & 53.5  \\
                        \cmidrule(lr){1-6}
                        \multirow{9}{*}{Ours} 
                         & \multirow{4}{*}{1} & head 1  & 41.0 & 56.7 & 44.8  \\
                         & & head 2  & 42.2 & 58.1 & 45.9  \\
                         & & head 3  & 41.8 & 57.8 & 45.6 \\
                         & & head 4  & 40.4 & 54.9 & 44.0 \\
                         \cmidrule(lr){2-6}
                         & \multirow{2}{*}{2} & head 1\&2 & 43.8 & 60.4 & 47.7  \\
                         & & head 2\&4 & 48.0 & 65.2 & 52.5 \\
                         \cmidrule(lr){2-6}
                         & \multirow{2}{*}{3} & head 1\&2\&3 & 49.7 & 67.5 & 54.4 \\
                         & & head 2\&3\&4 & 49.4 & 67.2 & 54.0 \\
                         \cmidrule(lr){2-6}
                         & 4 & head 1\&2\&3\&4 & 50.2 & 68.1 & 54.8  \\
                        \bottomrule[1.0pt]
                \end{tabular}
        }}
\caption{Diagnostic experiments on inquiry head combinations.}
\vspace{-2mm}
\label{tab:ablation-branch}
\end{table}

\subsubsection{Inquiry head combinations of \textit{\textbf{MI}}}
The above experiments have validated the effectiveness of \textbf{\textit{MI}} (\cref{tab:sota-combination}) and examined the influence of inquiry head number of \textbf{\textit{MI}} (\cref{tab:ablation-numbranch}). To experimentally explain why \textbf{\textit{MI}} is effective, we further conduct the experiments under different inquiry head combinations.  
Based on the trained model \#6 in \cref{tab:ablation-components}, we test the performance when enabling each inquiry head or multiple inquiry heads, and the results are reported in \cref{tab:ablation-branch}. 
Specifically, when enabling a certain inquiry head, the object queries from $k$-\textit{th} ($k$=1,2,3, or 4) inquiry head are used to predict final results. When enabling multiple inquiry heads (\eg, head 1\&2), the object queries from these inquiry heads are fused to predict final results.
When inquiry head number is 2 or 3, there are many inquiry head combinations, thus we present the results of two arbitrary combinations.

Overall, the performance of model configured with one inquiry head lags behind by significant margins, and fusing information in all inquiry heads achieves the best performance. Specifically, with the increasing of inquiry heads, the performance are gradually boosting. 
The reason is straightforward that one-time inquiry only enables object queries to learn the relatively limited information, which is similar to that each convolution kernel in CNNs only extracts the certain channel of feature.
Multi-time inquiries allow the model to fuse multi-pattern information learnt from different inquiry heads, though these information might be highly-collaborate (\eg, head 2 and 4) or weakly-collaborate (\eg, head 1 and 2).

\begin{figure}[h]
            \centering
            \includegraphics[width=0.48\textwidth]{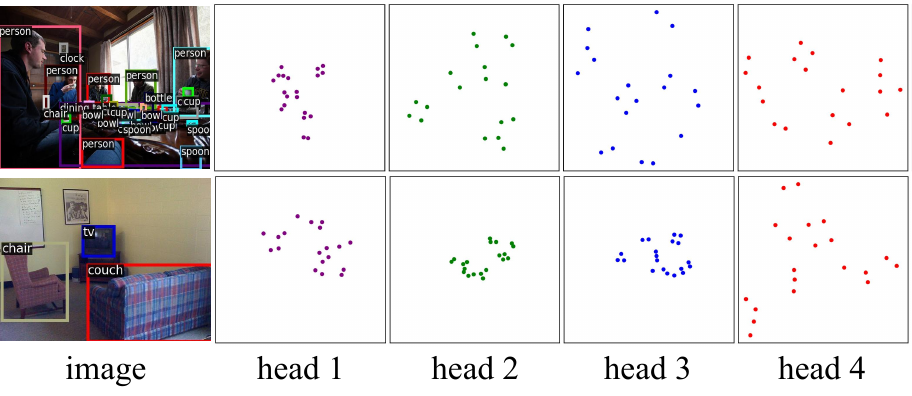}
            \caption{The visualization of object queries in different inquiry heads by T-SNE high-dimensional data visualization tool. \textbf{More results of \cref{fig:query_visual}, \cref{fig:head_visual}, and \cref{fig:result_visual} can be found in supplementary material.}
             }
            \label{fig:query_visual}
            \vspace{-2mm}
\end{figure}

\begin{figure*}[b!]
\vspace{-2mm}
            \centering
            \includegraphics[width=1.0\textwidth]{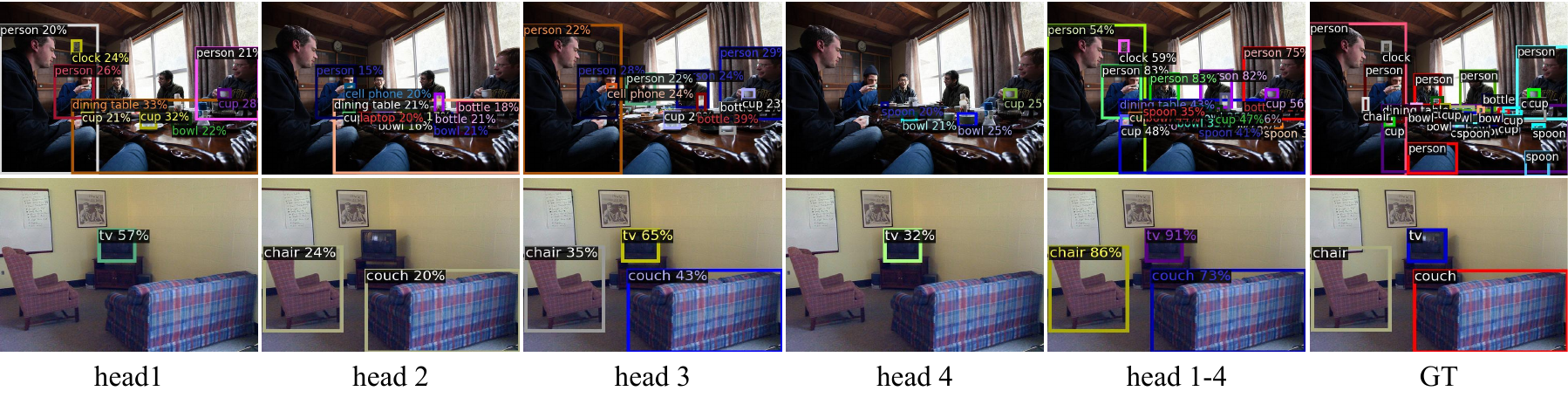}
            \caption{Object detection results based on the single inquiry head and multiple inquiry heads.
             }
            \label{fig:head_visual}
\end{figure*}

\subsubsection{Visualization analysis}
\label{sec:visualization}
The experiments in \cref{tab:ablation-numbranch} and \cref{tab:ablation-branch} have indicated that different inquiry heads are able to learn mutually collaborate information to improve object detection performance
when setting reasonable inquiry head number.
For deeper analysis, some visualization experiments are also conducted.\textbf{ Experiment 1 - object queries visualization:}
Object detection is performed on object queries at the last decoder layer,
thus visualizing object queries at the last decoder layer assists to distinguish what kind of information is finally learnt in individual inquiry head.
Therefore, we make use of the T-SNE tool to visualize the distributions of object queries at the last decoder layer.
Here, we would like to explain an object query, unlike CNN feature map, is a vector whose elements do not have the corresponding relations to image pixels, thus T-SNE visualization is adopted. Considering the majority of object queries represent the background and 93.51\% of images in COCO contain less than 20 objects, top-20 object queries are visualized in \cref{fig:query_visual}.
\textbf{ Experiment 2 - object detection results visualization:} If object detection results based on individual inquiry head could be visualized, the mutually collaborate mechanism of \textbf{\textit{MI}} could be straightly observed. Therefore, we visualize the object detection results using the single inquiry head and all inquiry heads as shown in \cref{fig:head_visual}.

We can observe from \cref{fig:query_visual} that object queries in different inquiry heads generally present distinct distributions, some are gathering and others are dispersing. Reflecting on detection results, different inquiry heads tend to focus on different kinds of information. For example, as shown
in \cref{fig:head_visual}, the head 2 seems to focus on big objects and head 4 intends to pay attention on small objects.
At the same time, some inquiry heads might focus on the similar pattern of information,\textit{ since sharing the information in the certain extent is the basis of collaboration}. For example, on the first example in \cref{fig:head_visual}, both head 1 and head 3 seem to struggle detecting objects in all regions of the image. This might partly explain the results in \cref{tab:ablation-numbranch} that continuously increasing inquiry heads will not constantly improve performance.
\textit{However, we note that, due to the diversity of images and ``black box'' characteristic of neural networks, it is challenging to clarify what specific patterns of information are learned through different inquiry heads, and it is also challenging to reflect the patterns constantly on diverse images.}

\begin{figure}[t]
            \centering
            \includegraphics[width=0.48\textwidth]{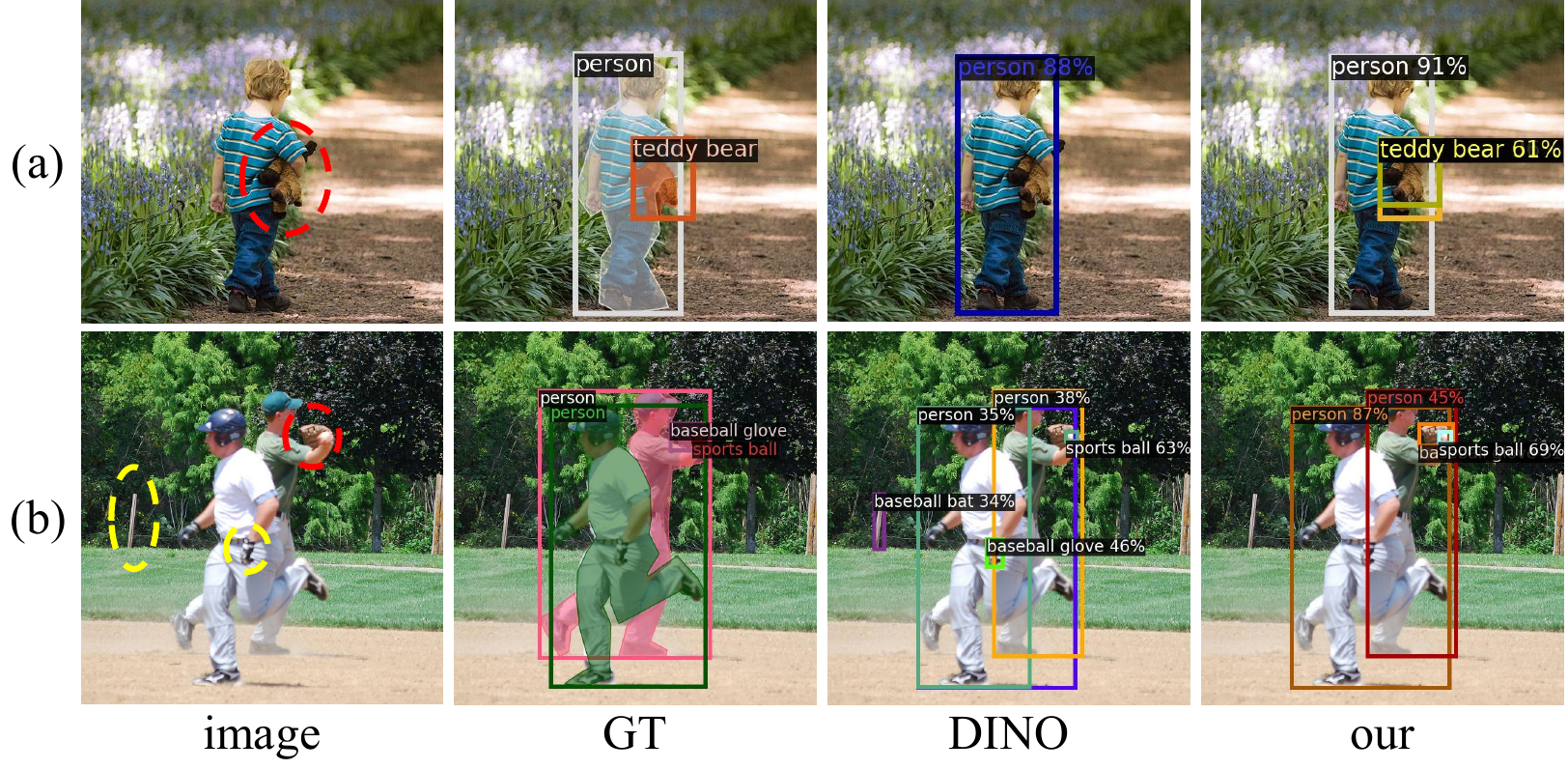}
            \caption{Qualitative comparison between DINO and our method. The red/yellow dashed circle indicates missing/false detection.
             }
            \label{fig:result_visual}
            \vspace{-2mm}
\end{figure}

We can also observe from \cref{fig:head_visual} that inquiry heads are mutually collaborate.
For example, on the first example, the ``clock'' is not detected in head 2, 3, and 4, and the ``cup'' in the hand of the leftest person is not detected in head 1, 2, 3, and 4, but these two objects are detected in head 1-4 and the confidences are improved. Thanks to the collaboration of multiple inquiry heads, our model exhibits stronger ability in detecting challenging objects.
For example, as shown in \cref{fig:result_visual}{\textcolor[RGB]{54,125,189}{a}}, the occluded ``teddy bear'' is successfully detected, while the baseline fails. Some false detections could also be avoided, an example is illustrated in \cref{fig:result_visual}{\textcolor[RGB]{54,125,189}{b}}.

\subsection{The Analysis of Complexity} 
The proposed \textbf{\textit{MI}} is implemented by parallelly adding decoder layers, which easily induce the misunderstanding that the performance improvement might result from increasing the parameters complexity. Therefore, to eliminate the misunderstanding, the models with different numbers of decoder layer and inquiry heads are tested, and the results are summarized in \cref{tab:ablation-layerbranch}. Note that we do not add \textbf{\textit{UFI}} (with 1M parameters) in the experiments to guarantee the fair comparison with the baseline and eliminate its disturbance on the testing.

By checking the results of DINO with 6, 12, and 24 decoder layers in \cref{tab:ablation-layerbranch}, we can observe that increasing the number of layers does not yield a performance improvement. Instead, a large amount parameters and computation are generated, even leading to performance degradation.
The potential reason is as follows. Due to the cascaded architecture of decoder layers, the representations update of object queries are constrained in the cascaded direction. When decoder layers are further increasing, the representations might be redundant or even negative, leading to performance degradation.

Compared to DINO with 12 decoder layers, our method with 6 decoder layers and 2 inquiry heads asks for less parameters and computation, and achieves better performance (49.5~v.s.~49.0).
Compared to DINO with 24 decoder layers, our method with 6 decoder layers and 4 inquiry heads also requires less parameters and presents better performance.
These results demonstrate our proposed \textbf{\textit{MI}} (rather than increasing computation complexity) contributes to the performance improvement. 

\begin{table}[t]
\centering
\scalebox{0.85}{
        \setlength{\tabcolsep}{2mm}{
        \renewcommand\arraystretch{1.0}
                \begin{tabular}{l|cc|ccc}
                        \toprule[1.0pt]
                        Method & LN & IHN & AP  & GFLOPS & Params\\
                        \midrule[1.0pt]
                        DINO    & 6  & 1  & 49.0 & 245 & 47M \\
                        \cmidrule(lr){1-6}
                        DINO    & 12 & 1  & 49.0 & 265 & 58M \\
                        Ours    & 6  & 2  & 49.5 & 263 & 57M \\
                        \cmidrule(lr){1-6}
                         DINO    & 24 & 1  & 47.3 & 301 &78M \\
                        Ours    & 6  & 4  & 49.8 & 299 & 75M \\
                        \bottomrule[1.0pt]
                \end{tabular}
        }}
\caption{Comparison on performance and complexity with the baseline. ``LN" and ``IHN" are short for ``Layer Number" and ``Inquiry Head Number", respectively.}
\vspace{-2mm}
\label{tab:ablation-layerbranch}
\end{table}

\section{Discussions}
\textbf{Cascade Multi-time Inquiries v.s. Parallel Multi-time Inquiries.}
The decoder layers of existing DETR-like models are actually performing cascaded multi-time inquiries. In contrast, our proposed \textit{\textbf{MI}} decoder layers adopt parallel multi-time inquiries. We would like to discuss the differences between them.

The essence of an inquiry head is object queries update their representations by interacting with image features.
As discussed in the introduction (\cref{sec:intro}), one-time inquiry is like that the student asks a question to the teacher, and obtains information about the image based on the teacher's answer.
For cascade multi-time inquiries, the student asks an initial question in the first inquiry head, and only asks the questions \textit{that are related with the initial question} in the following inquiry heads. Differently, {parallel multi-time inquiries} allow the student to simultaneously ask multiple questions from different views in the first decoder layer and constantly receive the gradually fine-grained answers for these questions in the following decoder layers.

\textbf{Real Parallel v.s. Pseudo Parallel.}
The number of object queries is tremendously larger than that of object GTs in an image. Therefore, the matching between GTs and object queries is an important issue that existing works are struggling to solve. The mainstream idea is setting up object queries for one-to-many matching ~\cite{chen2023group,jia2023detrs,zhao2024ms}.
However, to avoid NMS~\cite{neubeck2006efficient} post-processing operation, the object queries for one-to-one matching are also needed.
As a result, the object queries for one-to-one matching and the object queries for one-to-many matching are parallelly existing.
However, as explained in the introduction (\cref{sec:intro}), this parameters-sharing parallel architecture is a type of pseudo parallel. In contrast, our proposed parameters-dependent parallel architecture is real parallel, enabling to learn multi-pattern information. In addition, the query fusion mechanism makes multi-pattern information to be mutually collaborate and complementary.

\section{Conclusion}

This paper proposes a new decoder architecture, the core of which is the parallel \textbf{\textit{Multi-time Inquiries}} (\textbf{\textit{MI}}) mechanism. This mechanism presents the following advantages: \textit{\textbf{1)}} The mechanism improves feature utilization since it enables object queries to learn multiple patterns of information from image features; \textit{\textbf{2)}} The mechanism is simple and friendly to plug into existing DETR-like models; \textit{\textbf{3)}} The information learnt in different inquiry heads are mutually collaborate and complementary.

{
    \small
    \bibliographystyle{ieeenat_fullname}
    \bibliography{main}
}
\clearpage
\setcounter{page}{1}
\maketitlesupplementary

Due to the space limitation of the main text, we provide more results and discussions in the supplementary material, which are organized as follows:
\begin{itemize}

    \setlength{\itemindent}{0em}
    \item Section \ref{sec:supp_experiments}\textcolor[RGB]{54,125,189}{:} More Diagnostic Experiments.

    \setlength{\itemindent}{0em}
    -- Section \ref{sec:supp_effectiveness}\textcolor[RGB]{54,125,189}{:}  The effectiveness of our method on another DETR-like model.

    \setlength{\itemindent}{0em}
    -- Section \ref{sec:supp_lite}\textcolor[RGB]{54,125,189}{:}  The comparison between our proposed \textbf{MI-DETR} and \textit{Lite} \textbf{MI-DETR}.

    \setlength{\itemindent}{0em}
    -- Section \ref{sec:supp_fusion}\textcolor[RGB]{54,125,189}{:}  Diagnostic experiments on different query fusion mechanisms.

    \setlength{\itemindent}{0em}
    \item Section \ref{sec:supp_visualization}\textcolor[RGB]{54,125,189}{:}  More Visualization Analysis.

\end{itemize}

\section{More Diagnostic Experiments}
\label{sec:supp_experiments}
\subsection{The effectiveness of our method on another DETR-like model.}
\label{sec:supp_effectiveness}
We have conducted experiments to verify the effects of \textbf{\textit{MI}} on other models in \cref{experiments: combination}, including the most representative model DINO~\cite{zhang2022dino} and SOTA model Relation-DETR~\cite{hou2024relation}. Due to the space limitation of the main text, we present additional experiments based on recent proposed Align-DETR~\cite{cai2023align} to further validate the effectiveness and generalization of \textbf{\textit{MI}}.
The results are available in \cref{tab:supp_combination}, from which we can observe that our method show consistent improvement on Align-DETR.

\begin{table}[h]
\scalebox{0.9}{
        \setlength{\tabcolsep}{1.8mm}{
        \renewcommand\arraystretch{1.2}
                \begin{tabular}{llcll}
                        \toprule[1.0pt]
                        Method & Backbone & Epochs & AP  \\
                        \midrule[1.0pt]
                        Align-DETR~\cite{cai2023align}  & ResNet-50 & 12  & 50.2 \\
                        \rowcolor{lightgray!30} ours & ResNet-50 & 12  & \textbf{51.5(+1.3)} \\
                        Align-DETR~\cite{cai2023align}  & ResNet-50 & 24  & 51.3 \\
                        \rowcolor{lightgray!30} ours & ResNet-50 & 24  & \textbf{51.8(+0.5)} \\
                        \bottomrule[1.0pt]
                \end{tabular}
        }}
        \centering
\caption{Effectiveness of our method on Align-DETR.}
\vspace{-2mm}
\label{tab:supp_combination}
\end{table}

\begin{table}[h!]
\centering
\scalebox{0.9}{
        \setlength{\tabcolsep}{1mm}{
        \renewcommand\arraystretch{1.2}
                \begin{tabular}{c|ccc|cc}
                        \toprule[1.0pt]
                        Method & AP & $AP_{50}$ & $AP_{75}$ & GFLOPS & Params \\
                        \midrule[1.0pt]
                        \textbf{MI-DETR} & 50.2 & 68.1 & 54.8 & 311 & 76M \\
                        \textit{Lite} \textbf{MI-DETR} & 50.1 & 67.9 & 54.7 & 299 & 72M \\
                        \bottomrule[1.0pt]
                \end{tabular}
        }}
        \caption{The comparison between our proposed \textbf{MI-DETR} and \textit{Lite} \textbf{MI-DETR}.}
\label{tab:supp_MI}
\end{table}

\subsection{The comparison between our proposed \textbf{MI-DETR} and \textit{Lite} \textbf{MI-DETR}.}
\label{sec:supp_lite}
\textbf{MI-DETR} and \textit{Lite} \textbf{MI-DETR} are two kinds of architectures of our model, and the latter targets
to reduce the model complexity by sharing the self-attention layer in \textbf{\textit{MI}} decoder layers. As shown in \cref{tab:supp_MI}, \textit{Lite} \textbf{MI-DETR} achieves comparable performance with \textbf{MI-DETR} while requiring less parameters and computation.

\subsection{Diagnostic experiments on different query fusion mechanisms.}
\label{sec:supp_fusion}
To verify the impact of different query fusion mechanisms on model performance, we conduct experiments with three different fusion mechanisms, including ``add", ``linear+concat", and ``concat+linear". ``add" fusion directly adds up object queries from different inquiry heads. ``linear+concat" fusion first projects the object queries to $C/M$ dimensions along the feature dimension, and then concatenates them, where $C$ is the original dimensions of object queries and $M$ is the number of inquiry heads. ``concat+linear" fusion is the classic concatenation fusion as illustrated in \cref{eq:fusion}. The results are reported in \cref{tab:supp_fusion}, from which we can observe that the fusion mechanism has slight impact on the performance, which potentially proves that \textbf{\textit{MI}} is the main contributor to performance improvement.

\begin{table}[h!]
\centering
\scalebox{0.9}{
        \setlength{\tabcolsep}{1mm}{
        \renewcommand\arraystretch{1.2}
                \begin{tabular}{c|cccccc}
                        \toprule[1.0pt]
                        query fusion & AP & $AP_{50}$ & $AP_{75}$ & $AP_{S}$ & $AP_{M}$ & $AP_{L}$ \\
                        \midrule[1.0pt]
                        add            & 50.1 & 67.9 & 54.7 & 32.7 & 53.2 & 64.9 \\
                        linear+concat  & 49.8 & 67.7 & 54.2 & 32.7 & 52.9 & 64.8 \\
                        concat+linear  & \textbf{50.2} & \textbf{68.1} & \textbf{54.8} & \textbf{33.4} & \textbf{53.6} & \textbf{64.5} \\
                        \bottomrule[1.0pt]
                \end{tabular}
        }}
        \caption{Diagnostic experiments on different query fusion mechanisms in \textbf{\textit{MI}} decoder layer. ``add", ``linear+concat", and ``concat+linear" represent three different fusion mechanisms.}
        \vspace{-1mm}
\label{tab:supp_fusion}
\end{table}

\section{More Visualization Analysis}
\label{sec:supp_visualization}
\begin{figure*}[h]
            \centering
            \includegraphics[width=0.9\textwidth]{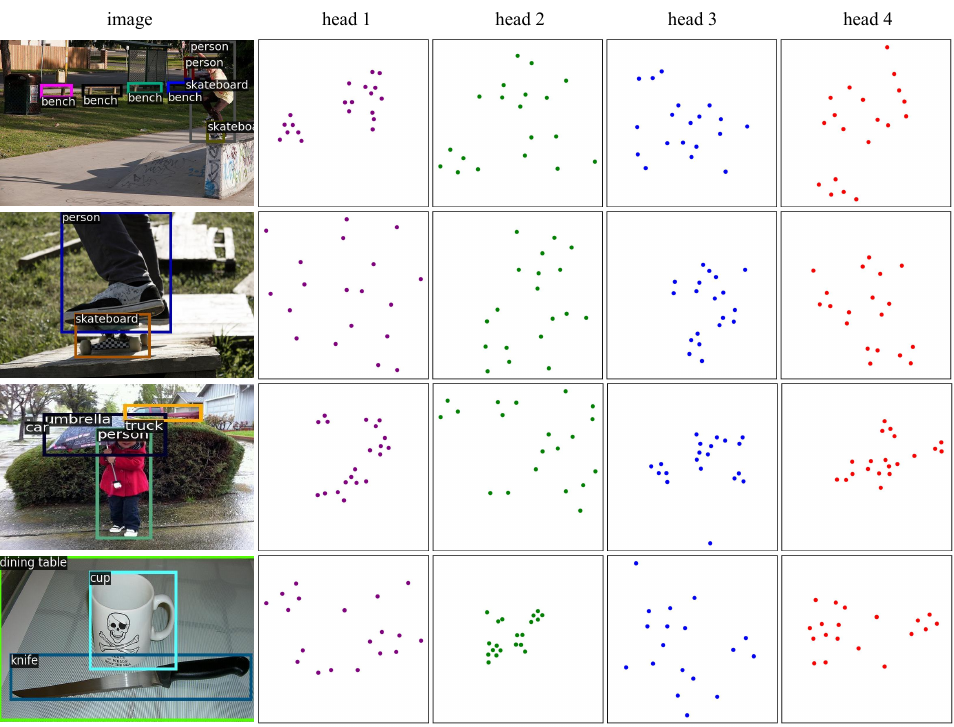}
            \caption{The visualization of object queries in different inquiry heads by T-SNE high-dimensional data visualization tool. This is a supplement to \cref{fig:query_visual} of the main text.
             }
            \label{fig:query_visual_more}
\end{figure*}

\begin{figure*}[h]
            \centering
            \includegraphics[width=0.9\textwidth]{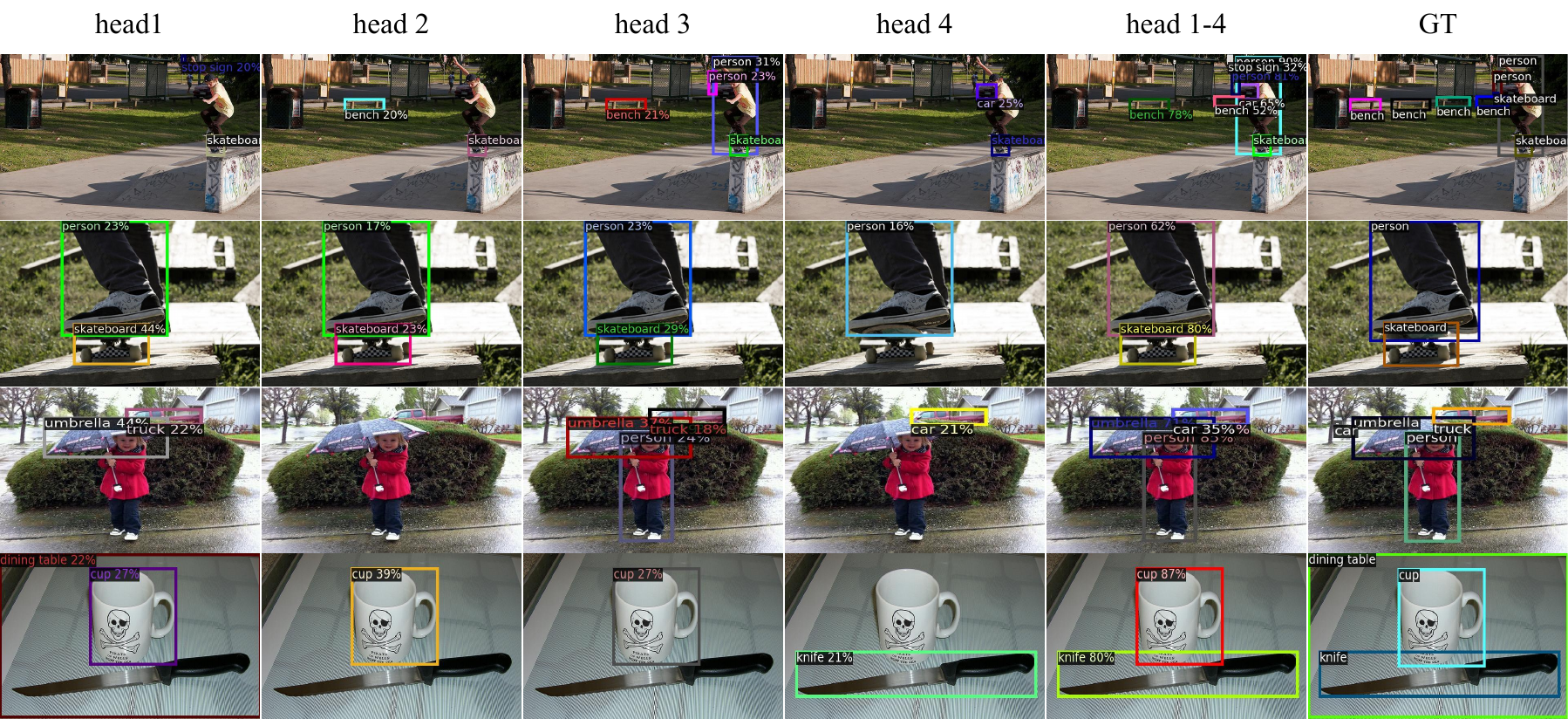}
            \caption{More object detection results based on the single inquiry head and multiple inquiry heads. This is a supplement to \cref{fig:head_visual} of the main text.
             }
            \label{fig:head_visual_more}
\end{figure*}

\begin{figure*}[h]
            \centering
            \includegraphics[width=1.0\textwidth]{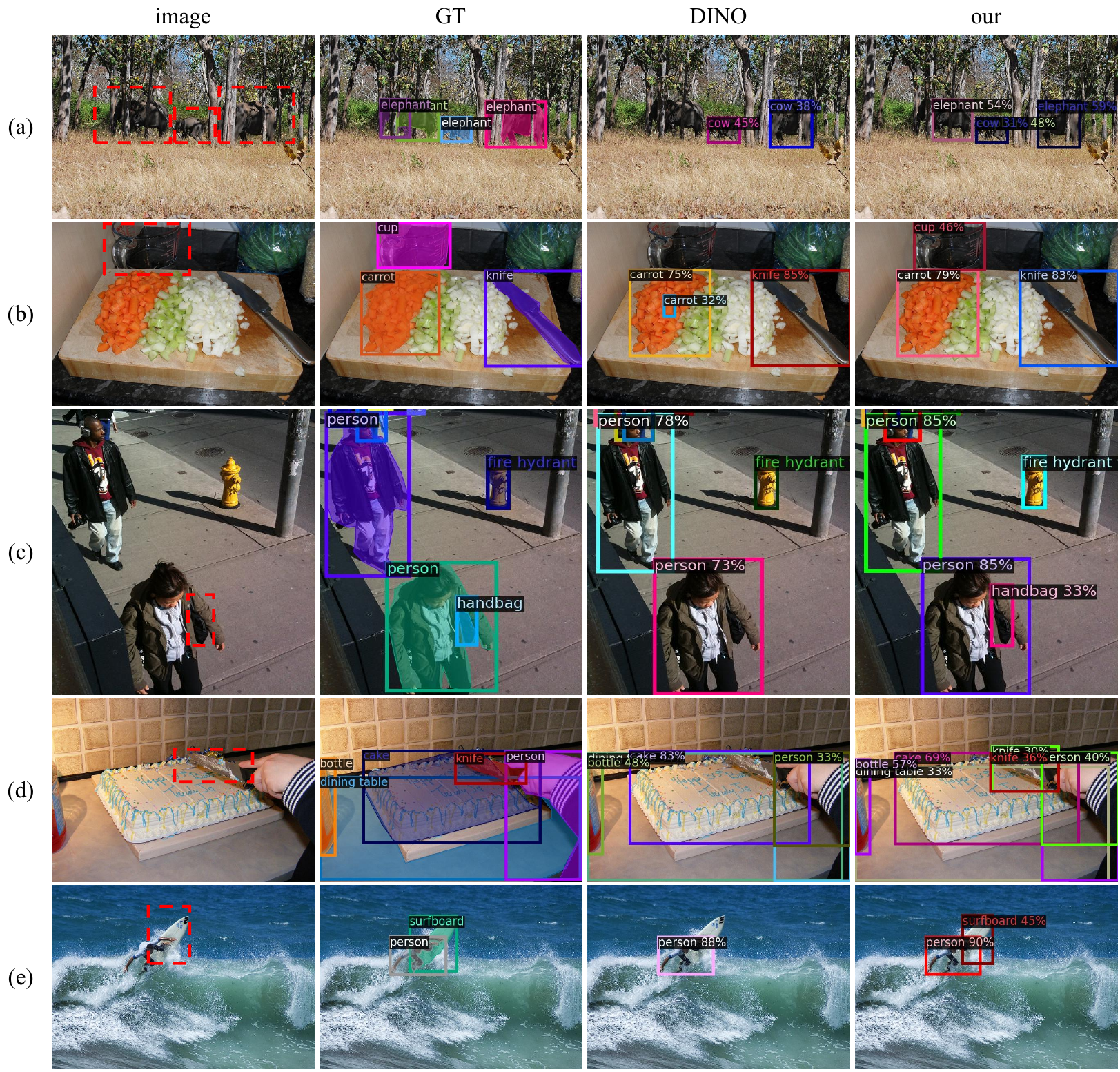}
            \caption{
            The ``elephants" in (a), the ``cup" in (b), and the ``surfboard" in (e) are confusingly mixed with the background. The ``elephants" in (a) and the ``handbag" in (c) are heavily-occluded. The ``handbag" in (c) and the ``knife" in (d) are small. These objects are difficult to detect, and DINO fails to detect them (\eg, the elephants are falsely detected as cows or missed). In contrast, our method successfully detect these challenging objects.
            Suggest zooming in to view clearer details. This is a supplement to \cref{fig:result_visual} of the main text.
             }
            \label{fig:result_visual_more}
\end{figure*}

We have conducted rich visualization experiments in \cref{sec:visualization}, including object queries visualization and object detection results visualization. To avoid the randomness of visualization results, we present more visualization results, as shown in \cref{fig:query_visual_more}, \cref{fig:head_visual_more}, and \cref{fig:result_visual_more}.
\cref{fig:query_visual_more} illustrates that object queries in different inquiry heads generally present distinct distributions, validating that multi-pattern information are learnt in different inquiry heads.
\cref{fig:head_visual_more} presents that different inquiry heads are mutually collaborate.
For example, on the first example, the ``car" is not detected in head 1, 2, and 3, and the ``person" on road is not detected in head 1, 2, 4.
These two objects are respectively noticed in head 4 and head 3 and are finally detected by heads 1-4.
As shown in \cref{fig:result_visual_more}, our method exhibits advantages in challenging natural scenes (\eg, extremely-small, heavily-occluded, and confusingly mixed with the background).

\end{document}